\documentclass[sigconf, screen, authorversion]{acmart}

\usepackage{siunitx}
\usepackage{subcaption}
\usepackage{graphicx}
\usepackage{caption}
\usepackage{multirow}
\usepackage[dvipsnames]{xcolor}
\usepackage{balance}
\usepackage{tabularx}
    \newcolumntype{C}{>{\centering\arraybackslash}X}

\AtBeginDocument{%
  }

\acmBooktitle{Proceedings of the 33rd ACM International Conference on Multimedia (MM '25), October 27--31, 2025, Dublin, Ireland}
\acmDOI{10.1145/3746027.3758297}
\acmISBN{979-8-4007-2035-2/2025/10}
\acmConference[MM '25]{Proceedings of the 33rd ACM International Conference on Multimedia}{October 27--31, 2025}{Dublin, Ireland}

\begin{document}

%%
%% The "title" command has an optional parameter,
%% allowing the author to define a "short title" to be used in page headers.
\title{OTR: Synthesizing Overlay Text Dataset for Text Removal}

%%
%% The "author" command and its associated commands are used to define
%% the authors and their affiliations.
%% Of note is the shared affiliation of the first two authors, and the
%% "authornote" and "authornotemark" commands
%% used to denote shared contribution to the research.
\author{Jan Zdenek}
\orcid{0009-0000-9761-8672}
\affiliation{%
  \institution{CyberAgent}
  \city{Tokyo}
  \country{Japan}
}

\author{Wataru Shimoda}
\orcid{0000-0001-6238-9697}
\affiliation{%
  \institution{CyberAgent}
  \city{Tokyo}
  \country{Japan}
}

\author{Kota Yamaguchi}
\orcid{0000-0002-3597-2913}
\affiliation{%
  \institution{CyberAgent}
  \city{Tokyo}
  \country{Japan}
}

%%
%% By default, the full list of authors will be used in the page
%% headers. Often, this list is too long, and will overlap
%% other information printed in the page headers. This command allows
%% the author to define a more concise list
%% of authors' names for this purpose.
\renewcommand{\shortauthors}{Jan Zdenek, Wataru Shimoda, and Kota Yamaguchi}

%%
%% The abstract is a short summary of the work to be presented in the
%% article.
\begin{abstract}
Text removal is a crucial task in computer vision with applications such as privacy preservation, image editing, and media reuse.
While existing research has primarily focused on scene text removal in natural images, limitations in current datasets hinder out-of-domain generalization or accurate evaluation.
In particular, widely used benchmarks such as SCUT-EnsText suffer from ground truth artifacts due to manual editing, overly simplistic text backgrounds, and evaluation metrics that do not capture the quality of generated results.
To address these issues, we introduce an approach to synthesizing a text removal benchmark applicable to domains other than scene texts.
Our dataset features text rendered on complex backgrounds using object-aware placement and vision-language model-generated content, ensuring clean ground truth and challenging text removal scenarios.
The dataset is available at \url{https://huggingface.co/datasets/cyberagent/OTR}.
\end{abstract}

%%
%% The code below is generated by the tool at http://dl.acm.org/ccs.cfm.
%% Please copy and paste the code instead of the example below.
%%



%%
%% Keywords. The author(s) should pick words that accurately describe
%% the work being presented. Separate the keywords with commas.
%% A "teaser" image appears between the author and affiliation
%% information and the body of the document, and typically spans the
%% page.
% \begin{teaserfigure}
%   \includegraphics[width=\textwidth]{sampleteaser}
%   \caption{Seattle Mariners at Spring Training, 2010.}
%   \Description{Enjoying the baseball game from the third-base
%   seats. Ichiro Suzuki preparing to bat.}
%   \label{fig:teaser}
% \end{teaserfigure}

% \received{20 February 2007}
% \received[revised]{12 March 2009}
% \received[accepted]{5 June 2009}

%%
%% This command processes the author and affiliation and title
%% information and builds the first part of the formatted document.

\maketitle

\section{Introduction}
% What is text removal and why is it important?
Text removal aims to erase texts from images and fill in the background with seamless pixels, preserving the visual quality of the images.
Text removal has a wide range of applications, such as removing captions from videos \cite{lee2003automatic, mosleh2013automatic}, obscuring private or sensitive information \cite{nakamura2017scene, liu2020erasenet, wang2023real, peng2024viteraser}, and editing text in images \cite{wu2019editing}, where removal is usually an important first pre-processing step for the following workflow.

% Domain generalization problem
The majority of text removal literature focused on the removal of text from natural scene images \cite{nakamura2017scene, zhang2019ensnet, liu2020erasenet, zdenek2020erasing, wang2023real, peng2024viteraser}, commonly referred to as scene text.
One of the early works on scene text removal introduced benchmark datasets \cite{zhang2019ensnet} that have remained the primary standard for evaluation up to the present, where scene examples often include traffic signs or billboards.
While scene texts are an important target domain, we argue that the current benchmark is not always applicable to evaluating text removal in other domains, such as creative domains involving printed posters or advertising banners, which present a different requirement in background inpainting.
For example, scene texts do not cross object boundaries, but poster texts can appear on top of a background containing multiple objects.

% Artifacts problem
Another qualitative problem in the existing benchmark is that the dataset presents pixel-level artifacts originating from manipulating ground truth images using an image editing tool, and sometimes leaves visible noise in the inpainted background.
These artifacts introduce noise in the evaluation metrics such as PSNR, but they are hard to fix because we cannot remove text in real world scenes to collect perfectly clean ground truth.

% Our approach
In this paper, we address the limitations of the current scene text benchmarks with a synthetic approach.
Considering the creative domain in mind, we synthetically create overlay text on images that initially contain no text, and use the composited images to train or evaluate text removal models.
This approach ensures that the ground truth remains completely artifact-free because the background images are always clean without pixel-level manipulation.
Also, our synthetic approach can control text placement using the location of objects and concepts in the image.
This allows us to create more challenging text removal scenarios by positioning text over regions with complex structures and textures.
In experiments, we show that existing benchmarks are limited in accurately comparing qualitatively and how our synthetic benchmark can provide better measurement capability.

%% posisble applications: automatic translation (magazines, newspaper, etc)

The main contributions of the newly proposed dataset, which we call OTR (\textbf{O}verlay \textbf{T}ext \textbf{R}emoval), are as follows:
\begin{itemize}
\item We present a synthetic approach to build a dataset to evaluate text removal methods, which artificially overlays texts on a complex background. Our synthetic approach guarantees an artifact-free background in the ground truth.
\item We empirically study the evaluation capability of the proposed dataset and show that our approach can better capture the qualitative characteristics compared to the existing scene text benchmarks.    
% \item We propose to use additional metrics for the text removal task to better evaluate the visual quality of generated results. % not necessary
\end{itemize}

% Preparing ground truth images without degrading the quality and accuracy of ground truth is impossible.
%Scene text is not the only type of text in images for which the ability to remove it is in high demand.
%Computer vision models are also widely used to remove text from advertisements and banners to reuse raster image data with modified text.
% While scene text has been the primary focus of text removal research, there is also strong demand for removing text from advertisements, banners, and other forms of printed media, where modifying raster images for reuse is common.
% Considering the need to evaluate text removal performance on ads and banners, and the fact that it is impossible to create a dataset for scene text removal benchmark without noise introduced by manual editing of images, we propose a new dataset for text removal evaluation that simulates overlay text in advertisements and printed media.

\section{Related Work}
\subsection{Text Removal Methods}
The first attempts at text removal targeted captions and subtitles in videos, employing spatial and temporal restoration techniques across consecutive frames \cite{lee2003automatic, mosleh2013automatic}.
Owing to the advancements in image generation and editing achieved by deep learning methods, the recent focus of research has shifted toward scene text removal, that is, the task of eliminating text from natural scene images.
Nakamura et al.~\cite{nakamura2017scene} were the first to apply a convolutional neural network to remove scene text from images. Further research emerged shortly, with several methods \cite{zhang2019ensnet, liu2020erasenet, tursun2020mtrnet++, zdenek2020erasing} leveraging the progress made by generative adversarial networks (GANs) in the field of image generation and editing~\cite{goodfellow2014generative, isola2017image}.
Recent methods have adopted new techniques and architectures such as attention mechanism \cite{lee2022surprisingly} and vision transformer \cite{peng2024viteraser}.

Text removal methods can be classified into two categories: (1) one-stage methods, which directly transform input images containing text into text-free outputs in a single step \cite{nakamura2017scene, zhang2019ensnet, liu2020erasenet, tursun2020mtrnet++, wang2021pert, liu2022don, peng2024viteraser}, and (2) two-stage methods, which first explicitly detect or estimate text regions and then inpaint only those areas \cite{zdenek2020erasing, tursun2019mtrnet, tang2021stroke, keserwani2021text, conrad2021two, lee2022surprisingly, wang2023real, du2023progressive}.

\subsection{Text Removal Benchmarks}
SCUT~\cite{zhang2019ensnet} is a popular benchmark dataset for studying text removal and includes two benchmark sets, each built in a different approach.

\paragraph*{Manually created benchmark}
\emph{SCUT-EnsText}~\cite{zhang2019ensnet} benchmark consists of pairs of scene text images and their manually edited counterparts, where the text has been removed using Adobe Photoshop.
The scene text images have been collected from other scene text datasets, namely ICDAR 2015~\cite{karatzas2015icdar} and MLT~\cite{nayef2017icdar2017}.
Lyu et al. recently introduced another dataset for scene text removal~\cite{lyu2023fetnet}, which is built in a similar approach.
While manually created ground truth examples look natural, there is an inherent drawback that the background often exhibits pixel-level artifacts.

\paragraph*{Synthetic benchmark}
\emph{SCUT-SynText}~\cite{zhang2019ensnet} benchmark contains synthetic scene text images generated with the SynthText \cite{gupta2016synthetic} algorithm, along with corresponding background images serving as text-free ground truth.
The synthetic approach completely avoids the background artifacts present in the image manipulation, but has a limitation in that the resulting images do not always look natural as a scene image.
As SynthText is designed for studying scene text detection, texts are placed on a relatively uniform background, such as the sky, the water surface, or the ground.
This is fine for training a scene text detection model, but it leaves an issue for text removal because texts never appear on a complex background and tend to yield only easy-to-inpaint examples.
This limitation becomes problematic for studying text removal in a non-scene image.

% SCUT-EnsText contains images of text in natural real scenes, collected from other scene text datasets, namely ICDAR 2015 \cite{karatzas2015icdar} and MLT \cite{nayef2017icdar2017}. 
% We describe the issues with SCUT-EnsText benchmark and evaluation in Section \ref{sec:benchmark-issues}.

% SCUT-SynText consists of images with synthetically created text that imitates scene text but looks far from realistic, often being placed in the sky, on water surfaces or on the ground.
% Both SCUT-EnsText and SCUT-SynText exhibit limitations that stem from the ground truth creation approach --- namely, manual editing in Photoshop --- and the inherent characteristics of scene text, which mostly appears on simple, homogeneous backgrounds.
% While SCUT-SynText does not have artifacts in the background, the results look far from realistic, with text often being placed in the sky, on the water surface, or on the ground, which could be fine for studying text detection approaches, but not necessarily appropriate for the text removal evaluation, as the texts always appear on a uniform background.
% These factors diminish their effectiveness as a text removal benchmark.
% A detailed discussion of these issues is provided in Section \ref{sec:benchmark-issues}.

\section{Pilot Study: Benchmark Issues} \label{sec:benchmark-issues}
In this section, we study problems in the existing benchmarks.

\subsection{Editing Artifacts}
Manually created ground truth tends to contain a substantial number of pixels surrounding the text that differ from those in the originals.
Ideally, only the pixels corresponding to text strokes would be altered in the ground truth.
However, due to the limitations of image manipulation, achieving such precision in an image editor makes it virtually impossible, and broader areas around the text strokes are often modified to produce visually plausible results. As a result, there are artifacts and signs of use of image editing tools that make the ground truth differ from the original.

%% add discussion about the source of artifacts (manipulation - brush, sponge)
%% not only shadow and lighting difference

\begin{figure}
    \centering
    \includegraphics[width=0.24\linewidth]{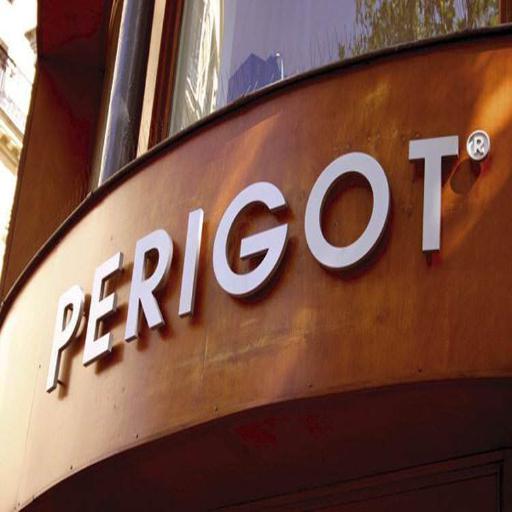}
    \includegraphics[width=0.24\linewidth]{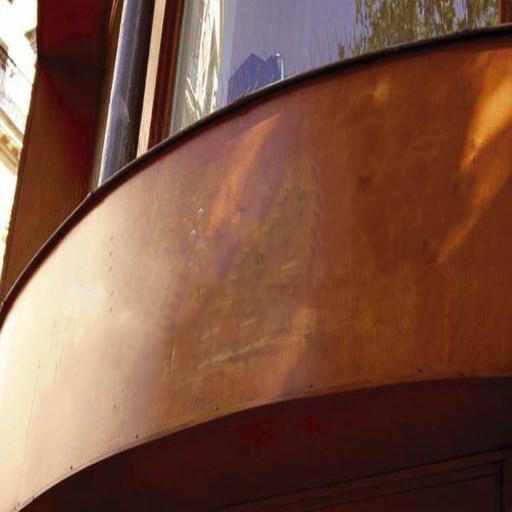}
    \includegraphics[width=0.24\linewidth]{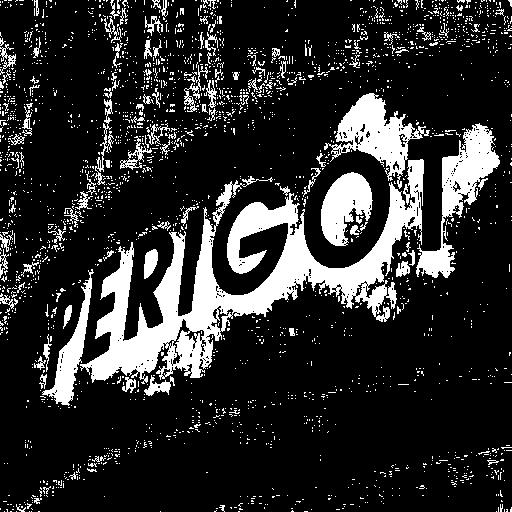}
    \includegraphics[width=0.24\linewidth]{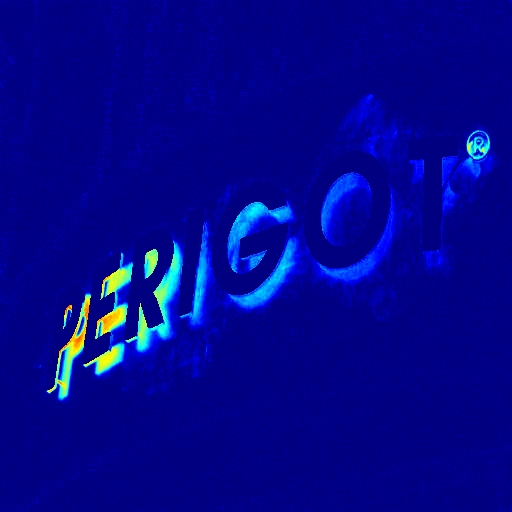}
    \includegraphics[width=0.24\linewidth]{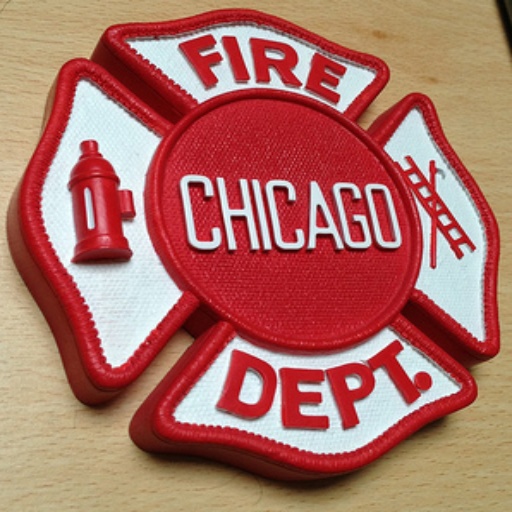}
    \includegraphics[width=0.24\linewidth]{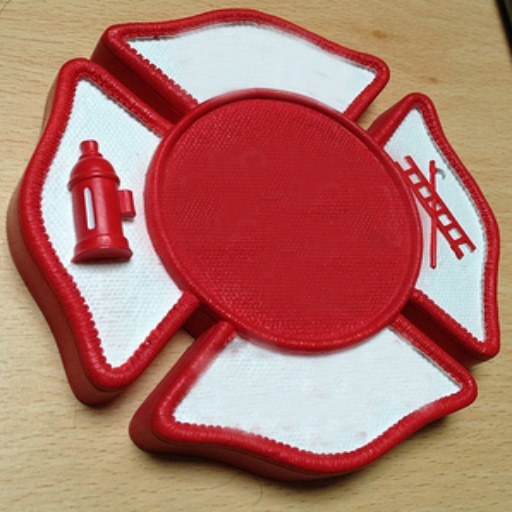}
    \includegraphics[width=0.24\linewidth]{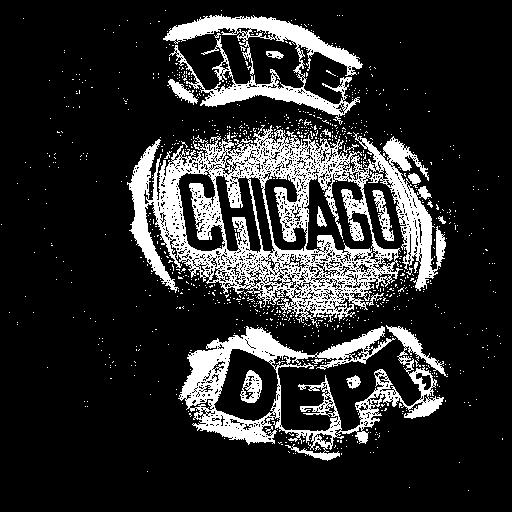}
    \includegraphics[width=0.24\linewidth]{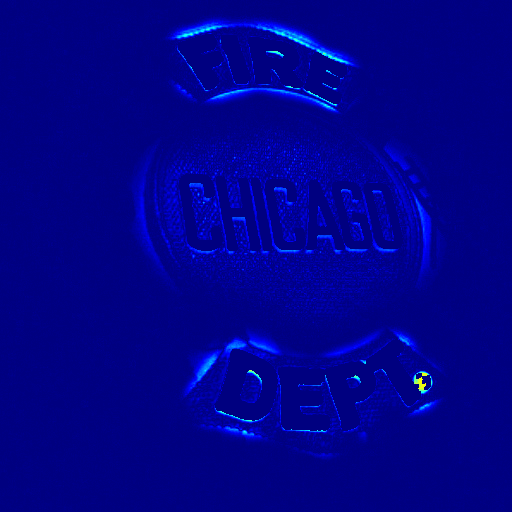}
    \caption{Examples from SCUT-EnsText dataset. From left to right, original image (first), ground truth image (second), map of pixels whose difference in value between the original and ground truth image exceeds the set threshold (third), the absolute difference between the original image and ground truth (fourth). The difference between the original and ground truth image should be as little as possible, but the surrounding regions contain altered pixels. Ideally, there should be no difference between the two when text stroke regions are excluded.}
    \label{fig:pixel_difference}
    \Description[Examples of artifacts caused by manual ground truth making.] {Illustrative examples from the SCUT-EnsText dataset. Each row shows a case study with four visualizations: the original image (left), its corresponding ground truth version with text removed (second), a binary difference map where pixel-wise differences between the original and ground truth exceed a defined threshold (third), and an absolute difference heatmap (right). These examples highlight the impact of text erasure and show that, ideally, the difference should be confined to the text stroke regions, with minimal alterations to surrounding background pixels.}
\end{figure}

We evaluate the discrepancy by PSNR between the original images and their corresponding ground truth in the SCUT-EnsText benchmark.
Figure \ref{fig:pixel_difference} shows examples of original and ground truth image pairs and their pixel-level differences.
Since the ground truth does not contain any text, we exclude text stroke regions from PSNR computation by detecting them with a text stroke segmentation model \cite{ye2024hi}.
We adopt a common implementation of PSNR that prevents division by zero by adding a constant $\epsilon = \num{1e-10}$ to the mean squared error (MSE). The PSNR between the original image $I$ consisting of $n$ pixels and its approximation $\hat{I}$ is then computed as follows:
\begin{equation}
    \text{PSNR} = 20 \cdot \log_{10}(m) - 10 \cdot \log_{10}(\text{MSE} + \epsilon),
\end{equation}
\begin{equation}
    \text{MSE} = \frac{1}{n} \sum_{i=1}^n (I_i - \hat{I}_i)^2.
\end{equation}
For 8-bit images with a maximum pixel value $m$ of 255, two perfectly identical images yield a PSNR of about 148 dB.
In contrast, the average PSNR across the whole test set of SCUT-EnsText is 42.72 dB, which roughly corresponds to $\frac{1}{10}$ of all pixels in the image differing by about 2.5\% in grayscale intensity.

To further quantify the discrepancies, we also compute the percentage of pixels --- excluding text stroke regions --- whose absolute difference in pixel value between the original image and ground truth exceeds a given threshold that we set to 3, a level sufficient to noticeably affect PSNR.
The results show that about 8\% of pixels exceed this threshold, indicating that a considerable number of pixels in the ground truth deviate from the original images.

In addition, since the SCUT-EnsText dataset is stored and distributed in JPEG format, there are additional compression artifacts that differ between the original images with text and their ground truth counterparts.
The visual artifacts --- originating both from manual editing and JPEG compression --- can compromise the fairness and reliability of evaluation results, particularly those that rely on pixel-level similarity between the original and ground truth.

\subsection{Uniform Background}
In real-world scenarios, scene text is typically placed on simple, uncluttered backgrounds to enhance its readability.
As a result, most text instances in the SCUT-EnsText dataset appear on plain, uniform backgrounds with minimal visual complexity, such as a billboard.
Similarly, in the case of SCUT-SynText, the SynthText algorithm --- designed to mimic geometrical properties of scene texts --- tends to place text in homogeneous regions with simple textures, such as sky, walls, or signboards.
As a consequence, the majority of text removal scenarios in both datasets is relatively easy, as it requires minimal understanding of structural context or semantic coherence.

In contrast, overlay text found in advertisements, magazines, and similar media can overlap with any part of the image, including complex objects.
This makes the text removal process more challenging, as it requires the model to preserve both structural integrity and semantic coherence in the inpainted regions.

We assess the complexity of text backgrounds in images using information entropy.
We first identify text stroke regions using a text stroke segmentation model \cite{ye2024hi}, then expand the detected text stroke regions $Z$ by applying dilation with a large kernel to obtain text vicinity regions $\tilde{Z}$.
The entropy of the background surrounding the text is then computed by:
\begin{equation} \label{eq:entropy-formula}
    H(X) = \sum_{x \in X} p(x) \log p(x),
\end{equation}
\begin{equation}
    X = I(\tilde{Z} - Z),
\end{equation}
where $I(\tilde{Z} - Z)$ denotes the region of the image that surrounds text stroke regions, selected by subtraction of masks $Z$ and $\tilde{Z}$.
As Table \ref{tab:complexity} in Section \ref{sec:dataset_complexity} shows, our new dataset exhibits higher entropy values, indicating that the backgrounds around text regions are more visually complex.

\subsection{Evaluation Approach}
\begin{figure}
    \centering
    \includegraphics[width=0.92\linewidth]{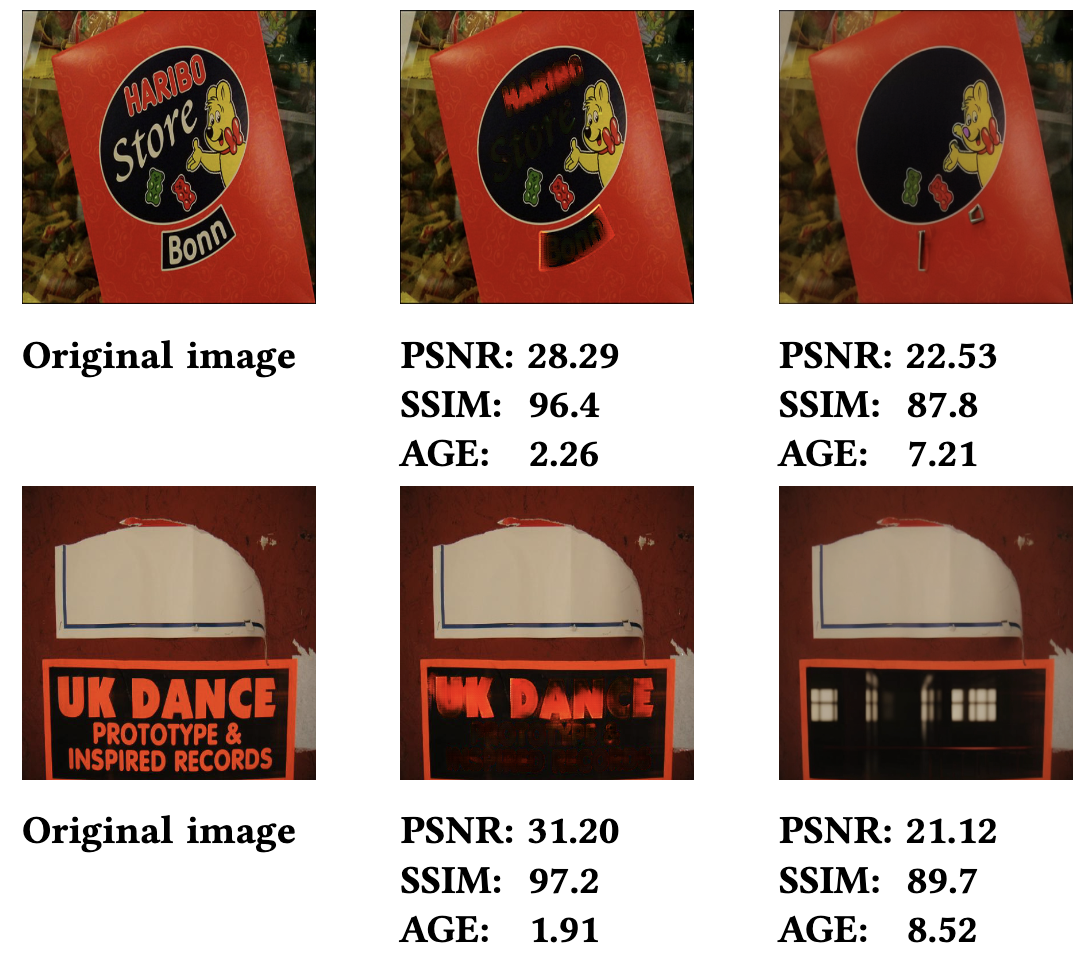}
    \caption{Results of text removal by EraseNet (middle) and FLUX.1 Fill (right). While PSNR, SSIM and AGE metrics suggest that results of EraseNet are better, results of FLUX.1 Fill look visually more convincing from a human perspective.}
    \label{fig:evaluation_comparison}
    \Description[Comparison of text removal results and misalignment of quantitative performance with human perception.]{Comparison of text removal results produced by EraseNet (middle column) and FLUX.1 Fill (right column) across two sample images. The original images are shown in the left column. Quantitative evaluation is provided below each image using PSNR (Peak Signal-to-Noise Ratio), SSIM (Structural Similarity Index), and AGE (Average Gray Error). Although EraseNet achieves better numerical performance across all metrics, the results of FLUX.1 Fill appear more visually coherent and natural from a human perceptual standpoint, highlighting a trade-off between metric-driven and perceptual quality assessments.}
\end{figure}

Commonly used metrics for evaluation of text removal --- such as structural similarity (SSIM), peak signal-to-noise ratio (PSNR), and the average of gray-level absolute difference (AGE) --- focus on how closely the output matches the ground truth. Other metrics such as text recall measure the amount of text remaining in the image, and Frechet Inception Distance (FID) measures the similarity between the original image and ground truth image distribution. However, text removal results can still be visually convincing even if they differ from the ground truth in the inpainted regions, especially since there is no single correct way to reconstruct the occluded content. Penalizing outputs that look natural but diverge from the ground truth overlooks this ambiguity. As illustrated in Figure \ref{fig:evaluation_comparison}, metrics commonly used for text removal evaluation often fail to account for this. Therefore, we advocate for the inclusion of evaluation methods that assess the visual quality of the results independently of their similarity to the ground truth.
\label{sec:inadequate_evaluation}

\begin{table}
\caption{Comparison of characteristics of different datasets.}
\begin{tabular}{l|c|c}
\hline             & artifact-free GT & complex text background  \\ 
\hline
SCUT-SynText &      \checkmark         &                          \\
SCUT-EnsText &                         &                          \\
OTR          &      \checkmark         &       \checkmark         \\
\hline
\end{tabular}
\label{tab:dataset-comparison}
\end{table}

Table \ref{tab:dataset-comparison} highlights the differences between individual datasets. SCUT-EnsText suffers from artifacts in the ground truth caused by manual editing during its creation. Additionally, since it features real-world scenes, most of the text appears on simple backgrounds that ensure good readability in real life. SCUT-SynText does not contain artifacts in the ground truth, but the design of the SynthText algorithm results in text being mostly placed on simple backgrounds. In contrast, OTR avoids the ground truth artifact issue by using synthetically generated images, and introduces more challenging text removal scenarios with text over complex backgrounds thanks to object-aware text placement.

\section{OTR Dataset}
This section describes the process used to construct our new dataset.

\subsection{Data Sources}
We use images and annotations from \textbf{Open Images V7} \cite{OpenImages} and \textbf{MS-COCO} \cite{lin2014microsoft} datasets to build our text removal dataset. Open Images V7 provides hierarchical class annotations with general and more fine-grained categories. From this dataset, we select images labeled with the following general classes: \textit{animal, food, furniture, home appliance, kitchen appliance, musical instrument, person, plant, sports equipment, tableware, toy, vehicle}. For MS-COCO, we rely on panoptic segmentation annotations and select images containing any of the following classes: \textit{dirt, floor, grass, pavement, river, road, sand, sea, sky, snow}. Both datasets are distributed under a CC BY 4.0 license, while individual images are licensed under CC BY 2.0.

\subsection{Dataset Creation}
We create paired data consisting of images with overlay text and their corresponding ground truth images without any text.
To ensure that the ground truth is clean and free from pre-existing text that we do not want to consider in our evaluation, we apply a scene text detection model \cite{liao2022real} to filter out any images that contain text prior to our processing.

In order to make text removal more challenging, we position the text in such a way that it overlaps with specific objects in the images. To achieve that, we use bounding box annotations of objects for images from Open Images V7 and place the text randomly within these regions.
For MS-COCO images, we utilize segmentation masks of our selected classes and place text randomly in the masked regions. All of our selected MS-COCO classes represent background and terrain elements, such as sky, sea, and road, which usually consist of simple textures that are easy to inpaint.

To render text on images, we use the skia-python \footnote{https://github.com/kyamagu/skia-python} graphics library and approximately 200 font files from Google Fonts \footnote{https://fonts.google.com/}. Font sizes are randomly selected from a predefined range, and long text is split into multiple text lines.
The text that we place into images is generated by a vision-language model (VLM) \cite{marafioti2025smolvlm} instructed to imagine an article or advertisement relevant to the image and make a short headline or a catchphrase for it. The exact prompt is:

\begin{quote}
    \textit{Think of a fictional article that is related to the image and think of a short phrase that could be a headline of the article, or think of a fictional advertisement that the image could be used for and think of a short phrase that could be used as a catchphrase for the advertisement. The phrase can be 1 to 20 words long.}
\end{quote}

We tried several VLMs, including PaliGemma 3B \cite{beyer2024paligemma} and CogVLM \cite{wang2024cogvlm}, and empirically found out that SmolVLM \cite{marafioti2025smolvlm} produces optimal results for our objective while also being efficient. In contrast, PaliGemma 3B generated very repetitive phrases, while CogVLM --- despite having eight times as many parameters as SmolVLM --- did not demonstrate any noticeable performance advantage.

\begin{figure}
    \centering
    \includegraphics[width=\linewidth]{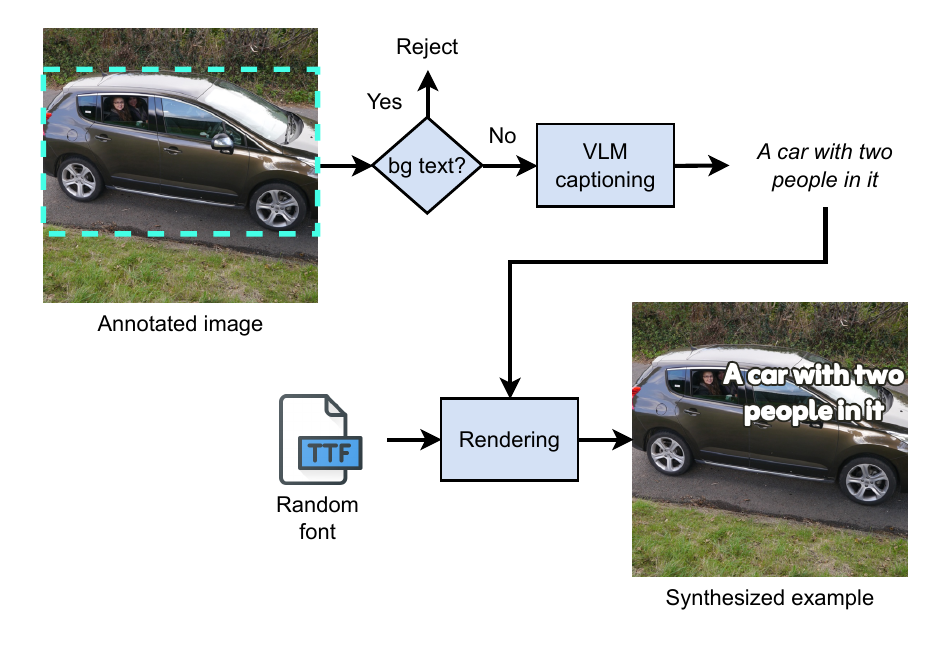}
    \caption{A diagram of our data generation process. We use a scene text detection model to filter out images that already contain text. Images with no detected text are passed to a VLM that generates short descriptive phrases, which are then rendered as overlay text on the images.}
    \label{fig:process}
    \Description[Overview of the synthetic data generation pipeline.]{Overview of the synthetic data generation pipeline. The process begins with annotated images, which are filtered using a scene text detection model to remove those containing pre-existing text. Images that pass this filter are sent to a vision-language model (VLM) that generates concise, descriptive captions. These captions are then rendered as overlaid text on the original images using randomly selected fonts, resulting in a synthesized dataset of captioned images.}
\end{figure}

Figure \ref{fig:process} provides an overview of our data generation process.

\subsection{Dataset Format}
The test set of our dataset consists of two subsets: \textit{OTR-easy}, containing data created from images sourced from the MS-COCO dataset, and \textit{OTR-hard}, containing data created from images sourced from Open Images V7. Tables \ref{tab:OTR-easy} and \ref{tab:OTR-hard} show the number of images per each class in both subsets, respectively. In total, \textit{OTR-easy} consists of 5,538 samples and \textit{OTR-hard} consists of 9,055 samples.

Each sample from the dataset consists of an image with rendered text, its corresponding original image without any text, and word-level annotations specifying the bounding boxes of their locations along with their transcriptions. The images are stored in PNG format to avoid degradation by JPEG compression artifacts, and annotations are stored in JSON files.

We also present a training set consisting of about 74,716 samples that can be used for training from scratch or finetuning pretrained models for overlay text removal.

\begin{table}
\caption{Number of images in the \textit{OTR-easy} set per each class.}
\begin{tabular}{ll|ll}
\hline
dirt     & 1000 & road & 608 \\
floor    & 1000 & sand & 77  \\
grass    & 282  & sea  & 98  \\
pavement & 1000 & sky  & 417 \\
river    & 1000 & snow & 56 \\
\hline
\end{tabular}
\label{tab:OTR-easy}
\end{table}

\begin{table}
\caption{Number of images in the \textit{OTR-hard} set per each class.}
\begin{tabular}{ll|ll}
\hline
animal             & 714  & person           & 1000 \\
food               & 1000 & plant            & 1000 \\
furniture          & 1000 & sports equipment & 1000 \\
home appliance     & 188  & tableware        & 1000 \\
kitchen appliance  & 168  & toy              & 596  \\
musical instrument & 389  & vehicle          & 1000 \\
\hline
\end{tabular}
\label{tab:OTR-hard}
\end{table}

\section{Experiments}

\begin{table*}
\caption{Evaluation results on the OTR-hard dataset.}
\begin{tabular}{llllllllll}
 & PSNR $\uparrow$ & SSIM $\uparrow$ & AGE $\downarrow$ & pEPs $\downarrow$ & pCEPs $\downarrow$ & QualiCLIP $\uparrow$ & TOPIQ $\uparrow$ & LIQE $\uparrow$ & HyperIQA $\uparrow$ \\ 
\hline
EraseNet \cite{liu2020erasenet} (TIP '20)  & 26.24 & 93.66 & 4.09 & 0.038 & 0.026 & 0.688 & 0.526 & 3.47 & 0.545 \\
MTRNet\texttt{++} \cite{tursun2019mtrnet} (CVIU '20)   & 28.32  & 94.85 & 2.30 & 0.032 & 0.019 & 0.676 & 0.563 & 3.59 & 0.575 \\
PERT \cite{du2023progressive} (CVIU '23)      & 26.97 & 94.18 & 2.95 & 0.038 & 0.026 & 0.688 & 0.547 & 3.47 & 0.559 \\
SAEN \cite{du2023modeling} (WACV '23)      & 26.11 & 93.92 & 3.52 & 0.039 & 0.026 & 0.671 & 0.549 & 3.51 & 0.559 \\
ViT-Eraser \cite{peng2024viteraser} (AAAI '24) & 29.56 & 95.51 & 2.29 & 0.027 & 0.017 & 0.696  & 0.542 & 3.45 & 0.554 \\ \hline
DBNet\texttt{++} + SAM + LaMa & 31.76 & 95.74 & 1.82 & 0.025 & 0.013 & 0.725 & 0.551 & 3.71 & 0.561 \\
DBNet\texttt{++} + SAM + FLUX.1 Fill & 31.18 & 95.42 & 1.99 & 0.027 & 0.014 & 0.726  & 0.557 & 3.78 & 0.569  \\ 
DBNet\texttt{++} + SAM + SD 1.5 & 30.02 & 94.56 & 2.31 & 0.032 & 0.018 & 0.728 & 0.566 & 3.79 & 0.576 \\ \hline
\end{tabular}
\label{tab:results_hard}
\end{table*}

\begin{table*}
\caption{Evaluation results on the OTR-easy dataset.}
\begin{tabular}{llllllllll}
 & PSNR $\uparrow$ & SSIM $\uparrow$ & AGE $\downarrow$ & pEPs $\downarrow$ & pCEPs $\downarrow$ & QualiCLIP $\uparrow$ & TOPIQ $\uparrow$ & LIQE $\uparrow$ & HyperIQA $\uparrow$ \\ 
\hline
EraseNet \cite{liu2020erasenet} (TIP '20) & 28.85 & 94.83 & 3.88 & 0.039 & 0.026 & 0.733 & 0.520 & 3.71 & 0.549 \\
MTRNet\texttt{++} \cite{tursun2019mtrnet} (CVIU '20)  & 31.92 & 95.39 & 2.33 & 0.032 & 0.019 & 0.728 & 0.535 & 3.78 & 0.558 \\
PERT \cite{du2023progressive} (CVIU '23)      & 33.00 & 94.95 & 2.79 & 0.035 & 0.024 & 0.732  & 0.527 & 3.69 & 0.551 \\
SAEN \cite{du2023modeling} (WACV '23)      & 29.43 & 94.76 & 3.45 & 0.036 & 0.024 & 0.738 & 0.528 & 3.71 & 0.555 \\
ViT-Eraser \cite{peng2024viteraser} (AAAI '24) & 32.61 & 95.95 & 2.25 & 0.026 & 0.016 & 0.759 & 0.530 & 3.72 & 0.553 \\ \hline
DBNet\texttt{++} + SAM + LaMa & 52.31 & 96.06 & 1.79 & 0.024 & 0.012 & 0.761 & 0.538 & 3.87 & 0.558 \\
DBNet\texttt{++} + SAM + FLUX.1 Fill & 51.51 & 95.85 & 1.93 & 0.026 & 0.013 & 0.765 & 0.540 & 3.92 & 0.560 \\ 
DBNet\texttt{++} + SAM + SD 1.5 & 51.18 & 95.21 & 2.18 & 0.029 & 0.015 & 0.763 & 0.547 & 3.92 & 0.566 \\ \hline
\end{tabular}
\label{tab:results_easy}
\end{table*}

We use two types of methods to obtain baseline results for our benchmark:
(1) existing text removal methods \cite{liu2020erasenet, tursun2020mtrnet++, du2023progressive, du2023modeling, peng2024viteraser}, and (2) general image inpainting models \cite{suvorov2022resolution, flux2024, rombach2022high} combined with a separate text detector \cite{liao2022real} and Segment Anything model \cite{kirillov2023segment}.

Existing methods for text removal are pretrained specifically on data for scene text removal. In contrast, general inpainting models have been trained on large-scale image datasets, enabling them to perform effectively across a wide range of image domains. General inpainting models are used along with text detection models which detect all text regions in the image that have to be inpainted. To minimize the area that needs to be inpainted, bounding boxes produced by the text detector are further refined using the Segment Anything \cite{kirillov2023segment} model to segment the text strokes within them.

\subsection{Evaluation Metrics}
As discussed in Section \ref{sec:inadequate_evaluation}, commonly used evaluation metrics for text removal are not enough to thoroughly evaluate if text removal results look natural or not. To supplement commonly used metrics that depend on the similarity between generated results and ground truth images, we employ additional metrics designed for no-reference image quality assessment (NR-IQA). Namely, QualiCLIP \cite{agnolucci2024qualityaware} (a CLIP-based self-supervised method trained on increasingly degraded images), LIQE \cite{zhang2023liqe} (a multitask learning method leveraging knowledge from other tasks), TOPIQ \cite{chen2024topiq} (a top-down method focusing on semantically important local distortions) and HyperIQA \cite{Su_2020_CVPR} (a content-aware self-adaptive classification network). These methods are designed to assess the perceptual quality of images in accordance with human subjective perception.

Besides the newly adopted metrics, we also use metrics widely used in existing works on text removal methods, i.e., PSNR (peak signal-to-noise ratio), SSIM (structural similarity), AGE (average of gray-level absolute difference), pEPs (percentage of error pixels), and pCEPs (percentage of four-connected neighbors error pixels).

\subsection{Quantitative Evaluation}

Tables \ref{tab:results_hard} and \ref{tab:results_easy} show the evaluation results for our \textit{OTR-hard} and \textit{OTR-easy} datasets, respectively. As can be seen, the scores on the \textit{OTR-hard} datasets are lower for all methods across most metrics, indicating that \textit{OTR-hard} presents more challenging scenarios.

\begin{figure}
    \centering
    \includegraphics[width=\linewidth]{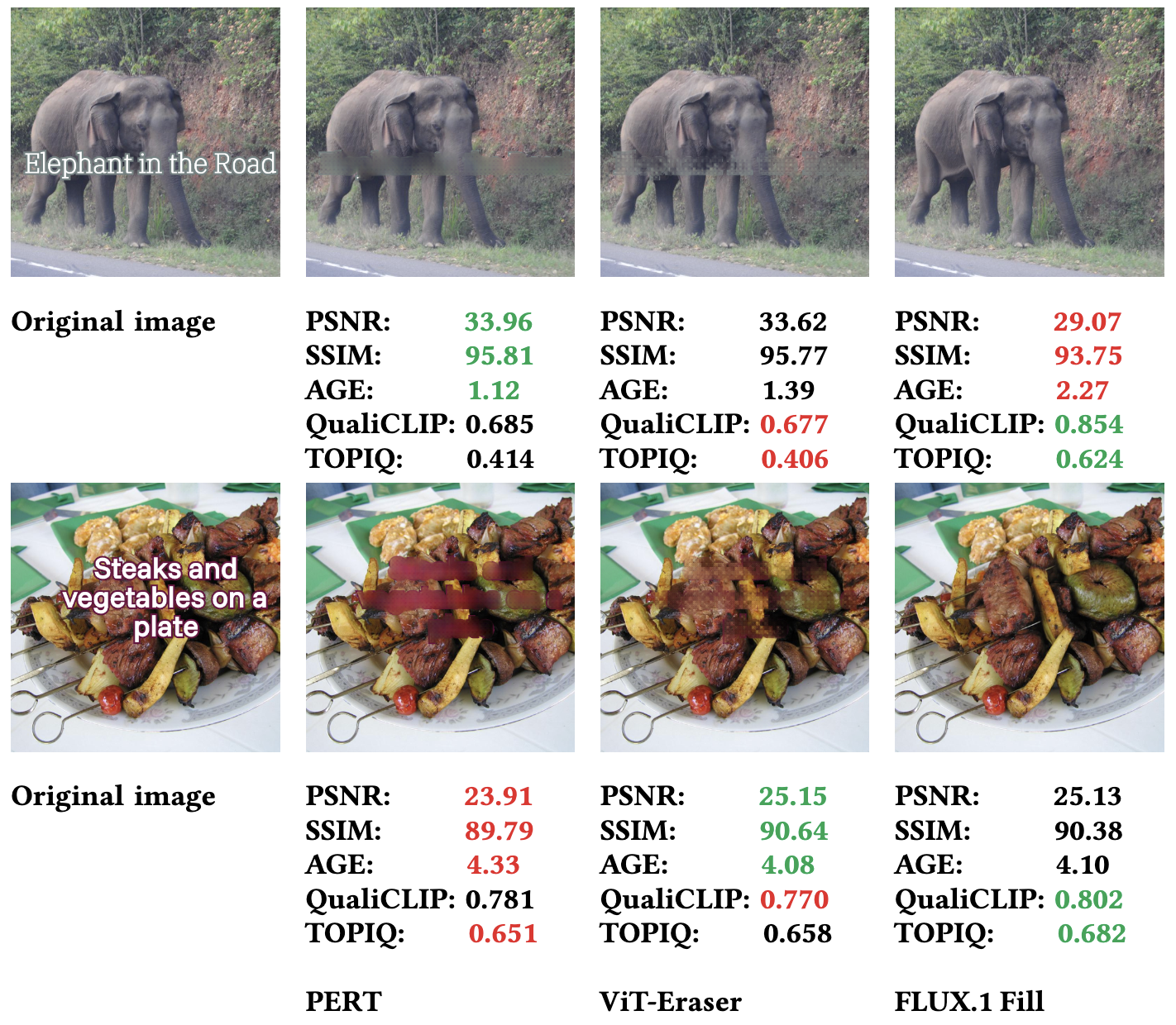}
    \caption{Text removal results produced by PERT, ViT-Eraser and FLUX.1 Fill and a comparison of discrepancy in results of different metrics.}
    \label{fig:metric_comparison}
    \Description[Quantitative and qualitative comparison of text removal results on our synthetic data.]{Quantitative and qualitative comparison of text removal results on our synthetic data produced by three methods: PERT, ViT-Eraser, and FLUX.1 Fill. Each row displays the original image followed by the outputs from each method, alongside numerical evaluation metrics including PSNR, SSIM, AGE, QualiCLIP, and TOPIQ. While PERT and ViT-Eraser exhibit higher PSNR and SSIM scores in some cases, FLUX.1 Fill achieves stronger performance in perceptual quality metrics (QualiCLIP and TOPIQ), indicating that it generates results that may be more visually pleasing from a human perspective.}
\end{figure}

\begin{figure*}[t!]
    \centering
    \begin{subfigure}[t]{0.3\textwidth}
        \centering
        \includegraphics[width=\linewidth]{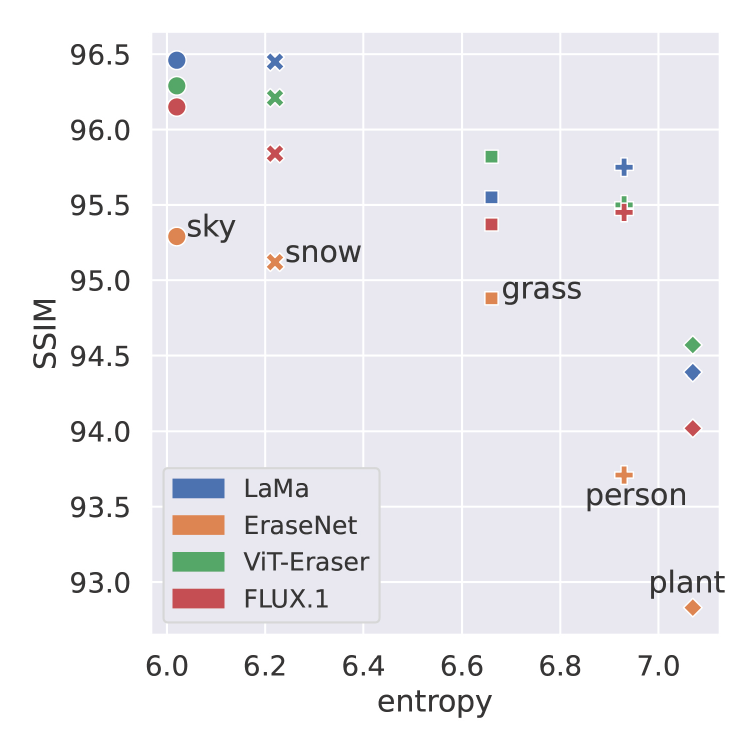}
    \end{subfigure}
    \begin{subfigure}[t]{0.3\textwidth}
        \centering
        \includegraphics[width=\linewidth]{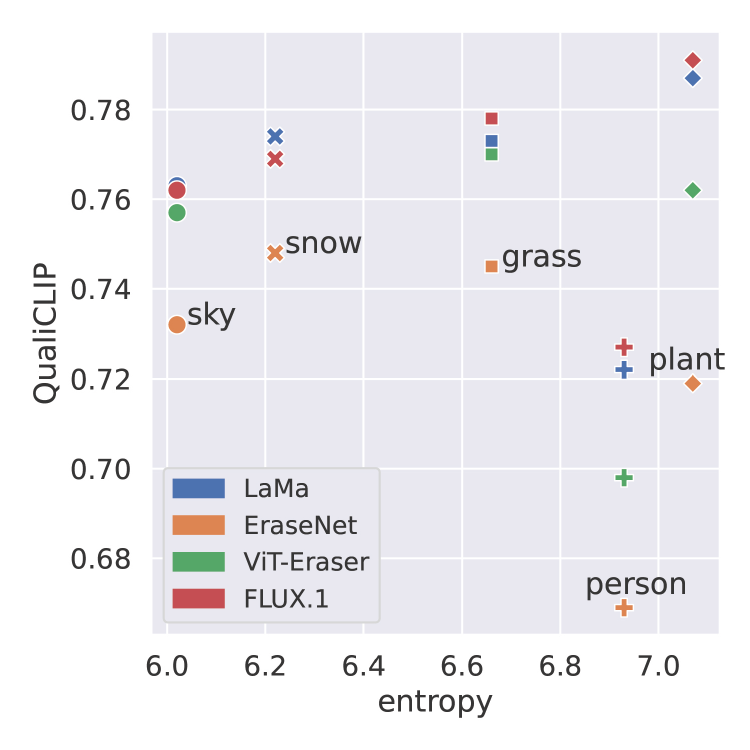}
    \end{subfigure}
    \begin{subfigure}[t]{0.3\textwidth}
        \centering
        \includegraphics[width=\linewidth]{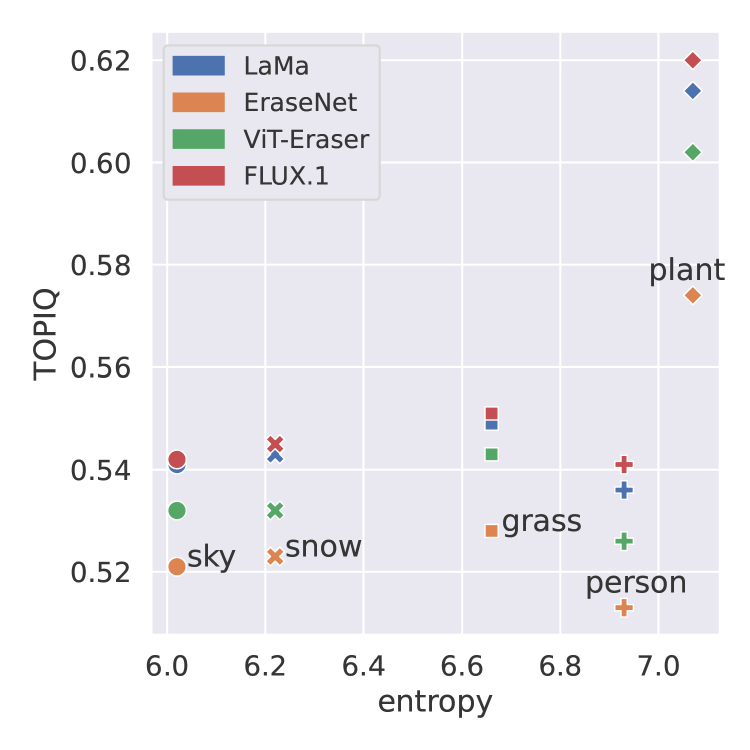}
    \end{subfigure}
    \caption{Correlation between the information entropy and several metric scores.}
    \label{fig:metric-entropy-correlation}
    \Description[Correlation plots showing the relationship between information entropy and three evaluation metrics - SSIM (left), QualiCLIP (middle), and TOPIQ (right)]{Correlation plots showing the relationship between information entropy and three evaluation metrics -SSIM (left), QualiCLIP (middle), and TOPIQ (right) - across different semantic categories including "sky", "snow", "grass", "plant", and "person". Each point represents a text removal result obtained using one of four methods: LaMa, EraseNet, ViT-Eraser, and FLUX.1. The analysis illustrates how increasing image entropy tends to affect metric scores differently}
\end{figure*}

\subsection{Significance of NR-IQA Metrics}

Figure \ref{fig:metric_comparison} shows text removal results using three different methods: PERT \cite{du2023progressive}, ViT-Eraser \cite{peng2024viteraser} and FLUX.1 Fill \cite{flux2024}. From a human perspective, FLUX.1 Fill produces better results, but as its outputs slightly differ from the ground truth, it underperforms in metrics relying on direct similarity to the ground truth. In contrast, NR-IQA metrics rank FLUX.1 Fill as the top-performing method.
This showcases a discrepancy between metrics focusing on the similarity between the result and ground truth and metrics focusing on perceptual image quality.
In practical use, visual quality often matters more than an exact match with the ground truth. This highlights the relevance of NR-IQA metrics for text removal evaluation.

\subsection{Dataset Complexity} \label{sec:dataset_complexity}

We use information entropy to measure the complexity of backgrounds around text as introduced in Equation \ref{eq:entropy-formula} in Section \ref{sec:inadequate_evaluation}. 

Table \ref{tab:entropy-hard} shows the information entropy for each class in \textit{OTR-hard}, the data set designed to feature text backgrounds that are more difficult to inpaint. In contrast, samples in \textit{OTR-easy} exhibit lower information entropy on average, as shown in Table \ref{tab:entropy-easy}. This indicates that text backgrounds in \textit{OTR-easy} are indeed less complex than those in \textit{OTR-hard}. Particularly, classes such as \textit{sky} and \textit{sea} yield the lowest entropy values, which aligns with our expectations since sea and sky are typically very simple and uniform textures.

Table \ref{tab:complexity} presents the information entropy of backgrounds around text instances across each dataset. Higher entropy values suggest that the backgrounds in our dataset, in particular the \textit{OTR-hard} set, are more complex, making it more challenging for text removal models to produce naturally looking results.

Figure \ref{fig:metric-entropy-correlation} illustrates the correlation between the information entropy and evaluation scores using the SSIM, QualiCLIP and TOPIQ metrics, respectively. The information entropy values correspond to several select classes from our dataset, namely \textit{sky, snow, grass, person} and \textit{plant}. For metrics that have been widely used for text removal evaluation, such as SSIM, performance on classes with lower information entropy tends to be better compared to those with higher entropy. This suggests a clear relationship between the quality of results and text background complexity, indicating that text on simple backgrounds is easier to remove.
Results of the QualiCLIP and TOPIQ metrics indicate that more advanced models that are better at understanding overall structure and semantics --- diffusion models such as FLUX.1 --- tend to outperform other methods particularly on classes with higher entropy values, i.e., classes with more complex backgrounds. This suggests that datasets featuring text on complex backgrounds are essential for highlighting the strengths of models with more advanced inpainting capabilities.
Additionally, the variation in metric scores across different methods is larger for classes with higher entropy, indicating that more challenging scenarios are more effective to distinguish the performance of text removal methods.

\begin{table}
\caption{Information entropy per each class in the \textit{OTR-hard} dataset.}
\centering
\begin{tabular}{ll|ll}
 \hline \multicolumn{4}{c}{ OTR-hard  ($H(X)$) } \\
\hline 
 animal & 6.91 & person & 6.93 \\
 food & 7.07 & plant & 7.07 \\
 furniture & 6.98 & sports equipment & 6.79 \\
 home appliance & 6.87 & tableware & 6.95 \\
 kitchen appliance & 7.02 & toy & 7.05 \\
 musical instrument & 6.89 & vehicle & 6.97 \\
\hline
\end{tabular}
\label{tab:entropy-hard}
\end{table}

\begin{table}
\caption{Information entropy per each class in the \textit{OTR-easy} dataset.}
\centering
\begin{tabular}{ll|ll}
 \hline \multicolumn{4}{c}{ OTR-easy  ($H(X)$) } \\
\hline
 dirt & 6.74 & road & 6.66 \\
 floor & 6.64 & sand & 6.60 \\
 grass & 6.66 & sea & 6.33 \\
 pavement & 6.77 & sky & 6.02 \\
 river & 6.70 & snow & 6.22 \\
\hline
\end{tabular}
\label{tab:entropy-easy}
\end{table}

\begin{table}
\caption{Comparison of text background complexity between individual datasets measured by information entropy.}
\begin{tabular}{ll}
\hline       & $H(X)$ \\ \hline
SCUT-EnsText & 6.32   \\
SCUT-SynText & 6.44 \\
OTR-easy     & 6.64   \\
OTR-hard     & 6.96 \\
\hline
\end{tabular}
\label{tab:complexity}
\end{table}

\section{Conclusion}
We introduced a new dataset for text removal that addresses key limitations of existing benchmarks, such as artifacts in ground truth and low background complexity. By simulating overlay text in advertisements and printed media, our dataset provides a more challenging and diverse testing benchmark. We also highlighted the need for better evaluation metrics that go beyond pixel-level similarity.

%%
%% The acknowledgments section is defined using the "acks" environment
%% (and NOT an unnumbered section). This ensures the proper
%% identification of the section in the article metadata, and the
%% consistent spelling of the heading.
% \begin{acks}
% To Robert, for the bagels and explaining CMYK and color spaces.
% \end{acks}

%%
%% The next two lines define the bibliography style to be used, and
%% the bibliography file.
\bibliographystyle{ACM-Reference-Format}
\balance
\bibliography{bibliography}

%%
%% If your work has an appendix, this is the place to put it.
% \appendix

\end{document}